\documentclass[conference]{IEEEtran}
\IEEEoverridecommandlockouts
\usepackage{cite}
\usepackage{amsmath,amssymb,amsfonts,nccmath}
\usepackage{graphicx}
\usepackage{textcomp}
\usepackage{xcolor}

\usepackage{bm}
\usepackage[noend]{algpseudocode}
\usepackage{algorithmicx}
\usepackage{algorithm}
\usepackage{multirow}
\usepackage{booktabs}
\usepackage{pifont}
\usepackage{verbatim}
\usepackage{balance}
\usepackage{footmisc}
\usepackage{verbatim}

\def\BibTeX{{\rm B\kern-.05em{\sc i\kern-.025em b}\kern-.08em
    T\kern-.1667em\lower.7ex\hbox{E}\kern-.125emX}}
\begin{document}

\title{FedSAE: A Novel Self-Adaptive Federated Learning Framework in Heterogeneous Systems
}
\author{\IEEEauthorblockN{Li Li\IEEEauthorrefmark{1}, 
		Moming Duan\IEEEauthorrefmark{1},
		Duo Liu\IEEEauthorrefmark{1},
		Yu Zhang\IEEEauthorrefmark{1},
		Ao Ren\IEEEauthorrefmark{1},
		Xianzhang Chen\IEEEauthorrefmark{1},
		Yujuan Tan\IEEEauthorrefmark{1},
		Chengliang Wang\IEEEauthorrefmark{1}
	}
	\IEEEauthorblockA{\IEEEauthorrefmark{1}College of Computer Science, Chongqing University, Chongqing, China\\
	}
	\IEEEauthorblockA{Email: $\{$li.li, duanmoming, liuduo$\}$@cqu.edu.cn
	}
}
\maketitle

\setlength\footnotemargin{0em}
\let\thefootnote\relax\footnotetext{ \rule[0.25\baselineskip]{0.5\columnwidth}{0.5pt}\\
	This paper will be presented at IJCNN 2021.\\
}

\begin{abstract}
Federated Learning (FL) is a novel distributed machine learning which allows thousands of edge devices to train model locally without uploading data concentrically to the server.
But since real federated settings are resource-constrained, FL is encountered with systems heterogeneity which causes a lot of stragglers directly and then leads to significantly accuracy reduction indirectly.
To solve the problems caused by systems heterogeneity, we introduce a novel self-adaptive federated framework \textit{FedSAE} which adjusts the training task of devices automatically and selects participants actively to alleviate the performance degradation. In this work, we

1) propose \textit{FedSAE} which leverages the complete information of devices' historical training tasks to predict the affordable training workloads for each device. In this way, \textit{FedSAE} can estimate the reliability of each device and self-adaptively adjust the amount of training load per client in each round.

2) combine our framework with Active Learning to self-adaptively select participants. 
Then the framework accelerates the convergence of the global model. In our framework, the server evaluates devices' value of training based on their training loss. Then the server selects those clients with bigger value for the global model to reduce communication overhead.

The experimental result indicates that in a highly heterogeneous system, \textit{FedSAE} converges faster than \textit{FedAvg}, the vanilla FL framework. Furthermore, \textit{FedSAE} outperforms than \textit{FedAvg} on several federated datasets --- \textit{FedSAE} improves test accuracy by $26.7\%$ and reduces stragglers by $90.3\%$ on average.

\end{abstract}


\section{Introduction}

Federated Learning (FL)~\cite{konevcny2016federated,mcmahan2017communication,smith2017federated}, famous for its significant contribution on privacy protection~\cite{yang2019federated}, is one of the most popular topics on distributed machine learning, widely used in personalized recommendation~\cite{ hard2018federated,yang2020federated} and clinical disease detection~\cite{brisimi2018federated}. 
Similar to the traditional distributed machine learning framework, FL performs most of its computation on remote clients but no data sharing between clients and the server in the FL training procedure.
Typical FL includes a central server and millions of remote clients such as personal computers, mobile phones, wearable devices, and so on. 
In each communication round, due to constraint network connectivity, the FL server only selects several participants which train a model such as neural networks locally, and then these participants upload their model parameters to the server periodically.
Finally, the server aggregates these uploaded model parameters to get a new global model. 
Usually, to train an FL model, several training rounds are performed, during which no raw data leaves clients.
This unique training mechanism keeps FL privacy-preserving compared with conventional distributed machine learning.

Nonetheless, a key challenge of FL is that the devices' configurations such as GPU, CPU, software, and network conditions are markedly different in real federated settings.
This challenge is called systems heterogeneity~\cite{smith2017federated,li2020federated} which causes the overhead of devices such as computation time and resources to accomplish the same task to vary heavily. 
Due to the different training overhead, some devices with high performance can complete their task. However, others called stragglers~\cite{li2018federated} with poor training conditions can only do partial training work, which leads to performance degradation of the global model.
Unfortunately, to handle the problem of stragglers, the FL server can only choose to wait or ignore them~\cite{bonawitz2019towards} because devices take full control of the connection at each round.
Systems heterogeneity heavily slows down the convergence rate and decreases the model accuracy~\cite{li2018federated} as the existence of numerous stragglers.

To solve the problem of systems heterogeneity, Nishio \textit{et al.}~\cite{nishio2019client} propose \textit{FedCS} which selects devices with enough resources to participate in FL training. 
However, this method requests the server to gather the available resource information of clients, which violates the principle of privacy protection. 
To alleviate the negative influence of stragglers, Li \textit{et al.}~\cite{li2018federated} propose \textit{FedProx} which allows each device to perform a variable amount of local work to endure the partial work of stragglers. 
\textit{FedProx} successfully promotes the test accuracy by 22\% on average compared with \textit{FedAvg}. 
However, \textit{FedProx} assumes that the partial work which stragglers can finish is known before stragglers drop out, which is impractical in a real federated setting. 
Moreover, the occurrence of drop out is unpredictable which means stragglers can not upload their partial work before they drop out. 
In most FL frameworks, the server usually assigns the same training workload to all of the selected clients, which neglects systems heterogeneity hence causes the high probability of stragglers.

In this paper, we propose \textit{FedSAE}, a self-adaptive FL framework aiming to alleviate the performance degradation caused by systems heterogeneity. 
In \textit{FedSAE}, the server self-adaptively adjusts the workloads allocated to clients according to their training history. 
Meanwhile, to accelerate the model convergence speed, we propose a mechanism based on active learning to select clients, which improves the quality of the global model.
We compare our framework with \textit{FedAvg}~\cite{mcmahan2017communication} in highly heterogeneous networks.
The experiment results convey that \textit{FedSAE} improves the absolute testing accuracy by 26.7\% and decreases the drop out rate by 90.3\% on average.

The main contributions of this paper are as follows:
  
\begin{itemize}
	
	\item We consider a new drop out scenario, which is more complex than the previous FL settings. In our work, the ability of clients to complete the training task is varied dynamically, so the probability of stragglers is also dynamic. Therefore, we propose a workload prediction algorithm for this heterogeneous scenario.

	\item We propose two federated algorithms, \textit{FedSAE-Ira} and \textit{FedSAE-Fassa}, to efficiently predict the affordable training workload of clients. Our methods are self-adaptive and privacy-preserving which works relying on the public training history of devices rather than any other private hardware information.

	\item We evaluate our methods with two models on four datasets. Comparing with baseline \textit{FedAvg}, our methods significantly upgrade the accuracy of the global model by 26.7\% and alleviate stragglers by 90.3\%.

\end{itemize}
  

\section{background and related work}\label{sec:background}

\subsection{Federated Learning}
In this subsection, we introduce the classical FL algorithm \textit{FedAvg}. And then, we briefly introduce systems heterogeneity.

\textbf{Federated Averaging.} 
Federated learning is introduced in~\cite{mcmahan2017communication} where McMahan \textit{et al.} propose the classical FL algorithm \textit{FedAvg}. 
In \textit{FedAvg}, the server coordinates clients to perform stochastic gradient descent (SGD) locally several iterations in parallel and updates global model by aggregating the uploaded model parameters from clients with Federated Averaging (\textit{FedAvg})\cite{mcmahan2017communication}. 
During the FL training process, no raw data but model weights are transmitted between clients and the server. 
Before \textit{FedAvg}, pioneers rely on full-batch SGD to train the federated learning model. 
To get an optimal model, the server usually training hundreds of rounds which produces a large communication overhead.
By utilizing the powerful computation ability of clients, \textit{FedAvg} forces clients to perform local mini-batch SGD several times to reduce the communication rounds and is communication-efficient. 

In \textit{FedAvg}, the particular object function is to optimize~\cite{li2018federated}:
\begin{equation}
	\min\limits_{{\bm{w}}} f(\bm{w}) = \sum_{k=1}^{N}\frac{n_k}{n} F_k(\bm{w}),
\end{equation}

Where $N$ is the number of total clients, $n$ is the total samples of $N$ clients i.e. $n=\sum_{k=1}^{N} n_k$, and $\frac{n_k}{n}$ is the weight of client $k$. For client $k$, the local objective function $F_k(\bm{w})$ is denoted by~\cite{li2018federated}:
\begin{equation}
    F_k(\bm{w}) = \frac{1}{n_k}\sum_{\bm{i} \in \mathcal{D}_k}f_i(\bm{w})
\end{equation}

where $\mathcal{D}_k$ is the training data on client $k$. 
$f_i(\bm{w})=\ell_i(x_i,y_i;w)$ is the loss function on examples $(\bm{x},\bm{y}) \in \mathcal{D}_k$. If $\mathcal{D}_k$ is sampled from global data uniformly, then $F_k(\bm{w}) \approx f(\bm{w})$ holds approximatively.
The approximation level between local objective and global objective improves as the increasing uniformity of client data $\mathcal{D}_k$.
That is why \textit{FedAvg} performs well on independently and identically distributed (IID) datasets.

In \textit{FedAvg}, at each round $t$, the server selects $K$ clients randomly ($K=C*N$, $C$ is the selected ratio in $N$ clients) and broadcasts the initial model parameter $w^t$ to participants.
The selected client (assumed client $k$) minimizes the local function loss $F_k(\bm{w})$ by performing SGD $E$ epochs locally (an epoch means training all the data on client $k$ once). 
At last, the server aggregates these parameters $w_k^{t+1}$ with federated averaging according to clients' weight $\frac{n_k}{n}$ to get the global parameter $w^{t+1}$ of round $t+1$. 
In general, the above process is executed $T$ rounds, i.e. the number of communication rounds is $T$. 
The details of \textit{FedAvg} is shown in Algorithm \ref{alg:fedavg}.

\begin{algorithm}[htbp]\small
	\caption{Federated Averaging (FedAvg)}
	\label{alg:fedavg}
	\begin{algorithmic}[1]
	    \Require initial model parameter $\bm{w}^0$, client fraction $C$, the number of clients $K$, the number of local epoch $E$, the learning rate $\eta$.
	    \Ensure the eventual model parameter $\bm{w}$.
	    \State Initialize $\bm{w}^0$.
	    \For{each communication round t = 1,2,3,...,T} 
	      \State $S_t$ $\leftarrow$ (Server selects $C*K$ clients clients randomly) 
	      \State Server broadcasts model parameters $\bm{w}_t$ to $S_t$.
	      \For{client $k \in S_t$ parallelly}
	        \State //clients perform SGD locally
	        \For{local epoch $e=1,2,3,...,E$}
	          \State $\bm{w}_{k}^{t+1} \leftarrow \bm{w}_{k}^{t} - \eta \nabla F_k(\bm{w}_k^t)$
	        \EndFor
	        \State $\bm{w}^{t+1} \leftarrow \sum_{k=1}^{K}\frac{n_k}{n}\bm{w}_{k}^{t+1}$
	      \EndFor
	    \EndFor
	    \State \Return $\bm{w} \leftarrow \bm{w}^{t+1}$
	    
	\end{algorithmic}
\end{algorithm}

\textbf{Systems heterogeneity.} Systems heterogeneity is a challenge in FL~\cite{smith2017federated}. 
It is inevitable because remote devices own varied configuration conditions such as GPU, CPU, memory, networks, etc. 
Due to systems heterogeneity, clients have different resource budgets thereby not all participants can finish their training tasks\cite{li2018federated}.
So some devices finish their training tasks fastly while some others called stragglers~\cite{li2018federated} train their model slowly even drop out.
These devices with better resource conditions could finish their training task more possible.
That is to say, systems heterogeneity causes heterogeneous training ability.
Conventional FL methods solve this problem by waiting for stragglers or omitting them\cite{mcmahan2017communication}, which leads to the accuracy degradation of the global model and slows down model convergence. 
Our work in this paper is to solve the systems heterogeneity challenges.

\subsection{Related Work}
\textbf{Federated Learning.} Based on the first work on FL algorithm \textit{FedAvg}~\cite{mcmahan2017communication}, 
the early FL studies are more about solving the statistical heterogeneity challenge in which the data distributed on devices is non-independent and non-identical (Non-IID). 
Zhao \textit{et al.}\cite{zhao2018federated} find that the accuracy of \textit{FedAvg} decreases by 55\% on highly skewed Non-IID datasets.
Thereby~\cite{huang2018loadaboost,zhao2018federated,duan2019astraea} aim to enhance the accuracy of the global model on Non-IID data. 
At the same time, that the global model converges slowly is another problem caused by Non-IID data\cite{li2018federated}, which induces \cite{li2019feddane,yao2019towards} to make efforts on increasing the convergence rate of the global model, \cite{yu2019parallel, li2018federated,li2019convergence} to provide convergence guarantees for Non-IID data.
Some pioneers work hard at reducing the communication overhead between clients and the server~\cite{konevcny2016federated,bonawitz2017practical,tang2018communication,samarakoon2018federated,caldas2018expanding,jeong2018communication}. 
For another aspect, some researchers try building a better FL system instead of optimizing the FL algorithm~\cite{mcmahan2017learning,bonawitz2017practical,bonawitz2019towards} where \cite{bonawitz2019towards} preliminarily explores a scalable production FL system for Android while \cite{mcmahan2017learning,bonawitz2017practical} provide FL more secure privacy protection.
However, the above works assume that all clients are equivalent in the training process, which causes a big performance variance in clients. 
Recently, some literature indicates that more fair local model performance distribution across clients contributes to a better global model~\cite{mohri2019agnostic,li2019fair}.
But the idea of performance fairness conflicts with privacy for the reason that it collects much sensitive information to ensure that the eventual distribution of local performance is fair enough. 

\textbf{Systems Heterogeneity and Straggling.} Systems heterogeneity generally exists in a real federated environment and it slows down the convergence rate of model~\cite{smith2017federated}.
Heterogeneous systems usually lead to the straggle issue in FL.
We can not eliminate systems heterogeneity but mitigate device straggling caused by it.
Some works reduce stragglers by selecting clients with enough resources~\cite{nishio2019client,rahman2020fedmccs}.
Nishio \textit{et al.}~\cite{nishio2019client} propose \textit{FedCS} which selects clients according to constrained time budget and resource budget to prevent participants from dropping out because of lack of resources~\cite{bonawitz2019towards}.
But it is difficult to judge whether the resource is enough or not.
Rahman \textit{et al.}~\cite{rahman2020fedmccs} propose another client selection framework \textit{FedMCCS} which selects clients according to multi-criteria such as location, resource, etc. 
However, the strict constraints of client selection may cause few participants thus slow down the convergence. 
Instead of selecting clients conditionally, Li \textit{et al.}~\cite{li2018federated} propose \textit{FedProx} which utilizes the partial work of stragglers to alleviate the model performance degradation. 
But \textit{FedProx} is too ideal to be implemented in real federated settings where whether a client will straggle is unknown.
To avoid wasting the partial work of stragglers,
Ferdinand \textit{et al.} \cite{ferdinand2020anytime} propose \textit{Anytime Minibatch} and Reisizadeh \textit{et al.} \cite{reisizadeh2019robust} propose \textit{QuanTimed-DSGD}. 
The two frameworks contribute to a novel training way that fixes the computation time of devices instead of fixing computation epochs.

\section{Federated Optimization: FedSAE}\label{sec:design}

In this section, we describe the motivation of this paper and then introduce the details of the proposed method \textit{FedSAE}.

\subsection{Motivation}\label{subsec:motivation}

Before introducing our framework, we conduct a simple experiment to show the motivation of this paper. 
To reveal the accuracy decrease of \textit{FedAvg} in heterogeneous systems, we implement a convex multinomial logistic regression task training with \textit{FedAvg} on MNIST~\cite{lecun1998gradient} dataset and Federated Extend MNIST (FEMNIST)~\cite{cohen2017emnist, caldas2018leaf} dataset.

For heterogeneous FL systems, a more reasonable design is to build a more complex federated environment, in which the client's ability to perform training tasks changes dynamically.
So we simulate a heterogeneous environment by varying the computing ability of each client with time separately. 
The affordable workload of each client is refreshed with a Gaussian Distribution in each round where the mean value $\mu$ is randomly initialized from the interval $[5,10)$, and the standard deviation $\sigma$ is randomly initialized from the interval $[0.25\mu,0.5\mu)$. 

\begin{figure}[t]
	\centering
	\includegraphics[]{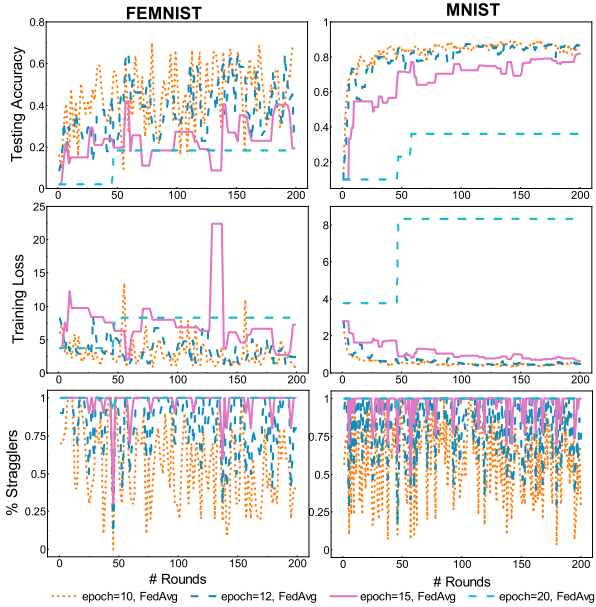}
	\vspace{-3mm}
	\caption{The results of \textit{FedAvg} trained on FEMNIST and MNIST datasets. The real affordable local epoch of each client is varying while the training epoch allocated to clients is fixed (i.e. $epoch=10,12,15,20$). With allocated epoch increasing from 10 to 20, averaging testing accuracy reduces by up to 53.3\% on MNIST, and the averaging drop out rate increases by up to 41\% on FEMNIST, which means considering systems heterogeneity, the performance of \textit{FedAvg} declines heavily. That's because many clients do not finish their assignments.}
	\label{fig:motivation}
\end{figure}

In our motivation experiment, we assign each client a varying amount of affordable local epoch each round, which quantifies the ability of clients to finish their training tasks. 
In the FEMNIST dataset, the number of clients is $N=200$, we set the number of selected clients $K=10$ per round. 
In MNIST dataset, $N=1000$, we set $K=30$.
We set learning rate $\eta=0.03$ to train $T=200$ rounds with \textit{FedAvg} and test the global model in each round. 
We run \textit{FedAvg} on FEMNIST and MNIST four times with $epoch=10, 12, 15, 20$ separately.
The top-1 testing accuracy of the global model, training loss, and drop out rate of clients are depicted in Fig.~\ref{fig:motivation}.

From Fig.~\ref{fig:motivation} we can see that taking the testing accuracy with $epoch=10$ as the reference accuracy, as the local epoch increases, the averaging testing accuracy on FEMNIST decreases by $5.2\%$ to $52.2\%$, and the average testing accuracy on MNIST decreases by $1.1\%$ to $53.3\%$. 
Also, we can see that as the local epoch increases, the training loss on the two datasets decreases more and more slowly. 
In extreme cases (i.e. $epoch=20$), the training loss even increases, which means that the training task fails because the model is diverging.
Besides, we find that the drop out rate fluctuates greatly when $epoch>10$, and almost all clients drop out when $epoch=20$. This is because \textit{FedAvg} cannot self-adapt the training task to the training ability of clients.
In short, Fig.~\ref{fig:motivation} shows that when FL systems are heterogeneous, the performance degradation of traditional FL framework (e.g. \textit{FedAvg}) is very obvious.
 
\subsection{Proposed Framework: FedSAE}\label{subsec:Proposed Frameowrk}
\begin{figure}[t]
	\centering
	\includegraphics[width=8.8cm]{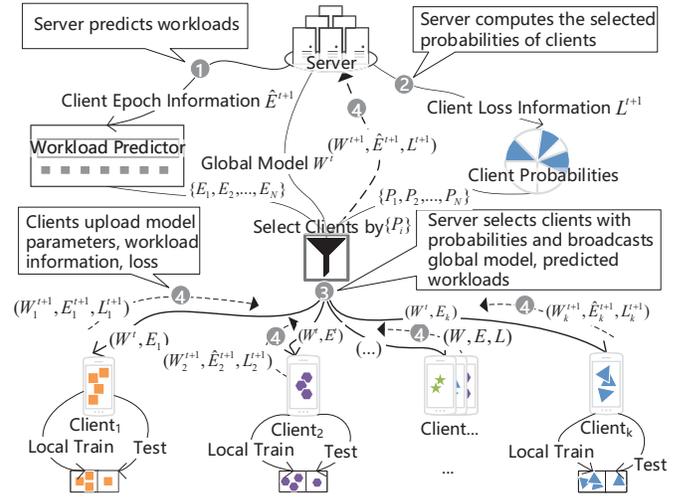}
	\caption{The workflow of \textit{FedSAE}.}
	\label{fig:FedSAE}
\end{figure}

To handle systems heterogeneity challenges in FL, we propose a new FL framework, \textit{FedSAE} which leverages workload prediction strategy based on the workload complete history of clients to predict the varied affordable workload of clients.
Besides, \textit{FedSAE} uses client selection strategy based on the training loss of clients to pick participants each round.
Our framework is similar to \textit{FedAvg} in the training process. Both of them include selecting clients, broadcasting models, locally performing SGD, uploading local models, and aggregating models. 
However, \textit{FedSAE} has two changes: predicting affordable workload and selecting participants with active learning (AL).
These changes make our framework more rational in practical federated settings. 
Before detailedly introducing our changes, we show our framework in Fig. \ref{fig:FedSAE}.

As shown in Fig. \ref{fig:FedSAE}, the training process of \textit{FedSAE} can be divided into four steps. First, the server predicts the affordable workload of clients according to their uploaded historical task completion information (\textcircled{1}). Second, the server converts clients' training loss into selected probability (\textcircled{2}). Third, the server picks participants according to probabilities in \textcircled{2} then broadcasts the global model and participants' predicted workload to participants (\textcircled{3}). At last, participants return model parameters, new task completion information and loss to the server and the server aggregates model weights for the next round (\textcircled{4}). We will introduce \textcircled{1} and \textcircled{2} in detail in the rest of this section.

\textbf{Affordable workload prediction.} 
As we discussed before, the ability of clients to perform training tasks in a heterogeneous system is varied dynamically. Clients participating in training without self-adaptively adjusting the number of training workloads are more likely to drop out, which leads to an inefficient global model. Therefore, it is necessary to predict the affordable workload of clients correctly. To better predict the affordable workload of clients, \textit{FedSAE} utilizes a task pair $(L_k^t, H_k^t)$ to limit the maximum and minimum values of affordable workload, where $L_k^t$ represents the amount of easy task, $H_k^t$ represents the amount of difficult task, obviously, $L_k^t \textless H_k^t$. Each client will maintain a task pair. The client first performs the easy workload $L_k^t$ and then increases its workload until it drops out or completes the difficult workload $H_k^t$.
This pair is beneficial in two aspects: 
(1) It makes sure that devices at least finish a conservative workload $L_k^t$ instead of dropping out.
(2) It predicts the affordable workload of clients to a specified range, which is more stable than directly accurate to the amount of affordable workload.

Considering the workload prediction task is similar to the congestion control task in communication technology, we get inspiration from classic TCP congestion control algorithms\cite{chiu1989analysis,he2010adaptive}. 
We propose two algorithms----\textit{FedSAE-Ira} (Algorithm \ref{alg:fedsae-ira}) and \textit{FedSAE-Fassa} (Algorithm \ref{alg:fedsae-fassa}) to predict the affordable workload of clients according to the historical task completion information.
For client $k$, it records $\tilde{E_k^t}$, the amount of workload it can accomplish recently, as the prediction basis of the workload prediction algorithms. 
Specifically, to predict the affordable workload of the next round, \textit{FedSAE-Ira} utilizes clients' task completion of the latest round while \textit{FedSAE-Fassa} utilizes clients' task completion of many past rounds.
The details of the two algorithms are as follows:

\begin{algorithm}[tbp]
    \setcounter{algorithm}{1}
	\caption{FedSAE-Ira: Federated Self-Adaptive Epoch with inverse ratio arise (Proposed Framework)}
	\label{alg:fedsae-ira}
	\begin{algorithmic}[1]
	\footnotesize
	    \Require the actually affordable local epoch $\tilde{E}_{k}^{t}$ of client $\bm{k}$ at round t, the server-predict workloads $\hat{E}_{k}^{t}$, the easy workloads $\hat{L}_{k}^{t}$, the difficult workloads $\hat{H}_{k}^{t}$.
	    \Ensure the predicted easy workloads $\hat{L}_{k}^{t+1}$,  the predicted difficult workloads $\hat{H}_{k}^{t+1}$.
		\Procedure {Server}{}
		\State Initialize $\bm{w}^0$, $\hat{L}_{k}^{0}$, $\hat{H}_{k}^{0}$
		\For{each communication round $t=1,2,...,T$}
			\State $S_t \leftarrow$ Select $K$ clients according to selected probabilities.
			\State Server broadcasts $\bm{w}_t$ to all selected clients.
			\State $( \hat{L}_{k}^{t}, \hat{H}_{k}^{t}, \tilde{E}_{k}^{t} ) \leftarrow$ Read history data from $S_t$.
			\For{each client $k \in S_t$ in parallel do}
			    \State $( \hat{L}_{k}^{t+1}, \hat{H}_{k}^{t+1}, \hat{E}_k^{t} ) \leftarrow$ \textbf{EpochPredict ($\hat{L}_{k}^{t}$, $\hat{H}_{k}^{t}$, $\tilde{E}_{k}^{t}$)}
				\State ${w}_k^{t+1}\leftarrow$~\textbf{Client($k$,~$\bm{w}^t$, $\hat{E}_k^{t}$)}
				
			\EndFor
			\State $\bm{w}^{t+1}\leftarrow \sum_{k=1}^{K} \frac{n_k}{n}\bm{w}_k^{t+1}$
		\EndFor
		\EndProcedure
		
		\Statex
		// run on server
		\Function{EpochPredict}{$\hat{L}_{k}^{t}$, $\hat{H}_{k}^{t}$, $\tilde{E}_{k}^{t}$}  
		
	// client $\bm{k}$ will not drop out.
		\If{ the actual affordable workloads greater than $\hat{H}_{k}^{t}$}
		    \State $\hat{L}_{k}^{t+1}, \hat{H}_{k}^{t+1} \leftarrow \hat{L}_{k}^{t} + \frac{\mathcal{U}}{\hat{L}_{k}^{t}}, \hat{H}_{k}^{t} + \frac{\mathcal{U}}{\hat{H}_{k}^{t}}$
		    \State $\hat{E}_{k}^{t} \leftarrow \hat{H}_{k}^{t}$
		
		// client $\bm{k}$ will drop out but weight at epoch $\hat{L}_{k}^{t}$ will be uploaded.
		\ElsIf{the actually affordable workloads fall into $[\hat{L}_{k}^{t}, \hat{H}_{k}^{t}]$}	    
		    \State $\hat{L}_{k}^{t+1} \leftarrow min(\hat{L}_{k}^{t} + \frac{\mathcal{U}}{\hat{L}_{k}^{t}}, \frac{1}{2}\hat{H}_{k}^{t})$
		    \State $\hat{H}_{k}^{t+1} \leftarrow max(\hat{L}_{k}^{t} + \frac{\mathcal{U}}{\hat{L}_{k}^{t}}, \frac{1}{2}\hat{H}_{k}^{t})$
		    \State $\hat{E}_{k}^{t} \leftarrow \hat{L}_{k}^{t} $ 
		    
		// client k will drop out and no weight will be uploaded.
		\Else
		    \State $\hat{L}_{k}^{t+1}, \hat{H}_{k}^{t+1} \leftarrow \frac{1}{2}\hat{L}_{k}^{t}, \frac{1}{2}\hat{H}_{k}^{t}$
		    
		    \State $\hat{E}_{k}^{t} \leftarrow 0$ 
		    
		\EndIf
		
		\State \Return $\hat{L}_{k}^{t+1}$, $\hat{H}_{k}^{t+1}$, $\hat{E}_k^{t}$ 
	    \EndFunction

		\Statex
		// run on client $\bm{k}$
		\Function{Client}{$k$,~$\bm{w^t}$, $\hat{E}_k^{t}$}

		\State $\mathcal{B} \leftarrow (seperate~\mathcal{P}_k~into~batches~of~size~B)$
				
		\For{each local epoch $e=1,2,...,\hat{E}_{k}^{t}$}
			\For{each local batch $b \in \mathcal{B}$}
				\State $\bm{w^t}\leftarrow\bm{w^t}-\eta\nabla L(b;\bm{w^t})$
			\EndFor
		\EndFor		
				
		\State \Return $\bm{w^t}$
		\EndFunction
	\end{algorithmic}
\end{algorithm}

\subsubsection{FedSAE-Ira}
\begin{figure}[t]
	\centering
	\includegraphics[width=8.8cm]{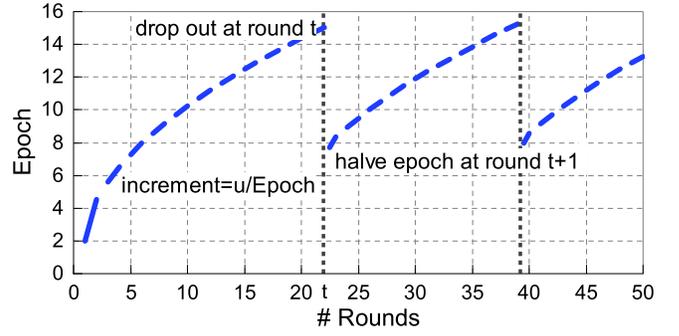}
	\vspace{-8mm}
	\caption{The simple process diagram of \textit{FedSAE-Ira}. If the client drops out at round $t$, the predicted workload of the next round $t+1$ turns to be halved. If not, it keeps increasing in inverse proportion e.g. client $k$ at $t$-th round increased by $\frac{\mathcal{U}}{E_k^{t}}$.}
	\label{fig:Ira-TCP}
\end{figure}

In Algorithm \ref{alg:fedsae-ira}, we show the simple process of \textit{FedSAE-Ira} in which the increased workload is inversely proportional to itself. 
If a client drops out, the client's predicted workload of the next round will be half of the current round otherwise the predicted workload of the client will increase by increment $\frac{\mathcal{U}}{\hat{E}_k^{t}}$ (see the line 11 to line 22 in Algorithm \ref{alg:fedsae-ira}).
As shown in Fig. \ref{fig:Ira-TCP}, at the beginning of the training the workload of clients is conservative and easy. Therefore, the workload increases fastly by increment $\frac{\mathcal{U}}{\hat{E}_k^{t}}$ which is inversely proportional to the amount of current workload until the client drops out in round $t$. And then, the workload of the client in round $t+1$ turns to half of the current workload. In this process, \textit{FedSAE-Ira} shows the following advantages: (1) It is cautious to increase the workload of clients in a gradually smaller increment, which can reduce the drop out probability of clients to a certain extent.
(2) It reduces workload exponentially after dropping out, which makes the workload recover to a safe level quickly. In this way, the client avoids dropping out consecutively.
We simply summarize the predicted workload of clients as the following equation:
\begin{equation}    
    \begin{split}
    \hat{E}_k^{t+1}= \left \{ 
    \begin{array}{ll} 
        \frac{1}{2}\hat{E}_k^{t},  & if~client~k~drops~out\\ 
        \hat{E}_k^{t}+\frac{\mathcal{U}}{\hat{E}_k^{t}},     & otherwise
    \end{array}
    \right.
    \end{split}
    \label{eqa:Ira}
\end{equation} 
where $\hat{E}_k^{t}$ is the predicted workload of client $k$ in $t$-th round, $\mathcal{U}$ is the hyperparameter to control the increment. 
An appropriate $\mathcal{U}$ is efficient because an excessively high $\mathcal{U}$ makes the prediction process fluctuated while an excessively low $\mathcal{U}$ causes the time converging to the affordable workload interval too long.
In \textit{FedSAE-Ira}, $L_k^t$ and $H_k^t$ are predicted according to~\eqref{eqa:Ira}. 
This method is a kind of Additive Increase Multiplicative Decrease (AIMD) method\cite{allman1999tcp} which is a feedback control algorithm with linear growth and exponential reduction. Chiu \textit{et al.}\cite{chiu1989analysis} demonstrate that for an additive increase and multiplicative decrease iterative process, the method is convergent.

\subsubsection{FedSAE-Fassa}
In order to make full use of the history of task completion, we propose \textit{FedSAE-Fassa} which utilizes the amount of the affordable workload of many past rounds to self-adaptively predict the affordable workload of the next round. 
As shown in Algorithm \ref{alg:fedsae-fassa}, first, the server calculates the workload threshold $\theta_{k}^t$ of $k$-th client in round $t$ as~\eqref{eqa:theta} based on the historical workload records (see the line 8 in Algorithm \ref{alg:fedsae-fassa}). $\theta$ represents the number of workloads that the client can complete in most cases.
\begin{equation}
    \theta_{k}^{t+1}=\alpha\theta_{k}^{t}+(1-\alpha)\tilde{E}_{k}^{t}
    \label{eqa:theta}
\end{equation}
Where $\tilde{E}_{k}^{t}$ is the real affordable workload, $\alpha$ is the smoothness index which decides the percentage of the affordable workload of the last few rounds in threshold $\theta_{k}^{t}$. For example, a small $\alpha$ means that the workload threshold $\theta_{k}^{t}$ is biased towards $\tilde{E}_{k}^{t}$, the affordable workload of the latest round.  
Equation \eqref{eqa:theta} is a simple expression of Exponential Moving Average (EMA)\cite{haynes2012exponential} which means in $\theta_{k}^{t}$ the weights of old affordable workload (e.g. the epochs before round $t$) decrease exponentially as the training rounds increase. After obtaining $\theta_{k}^t$, the clients get predicted workloads of round $t+1$ according to the following equation:
\begin{equation}    
    \begin{split}
    \hat{E}_k^{t+1}= \left \{ 
    \begin{array}{ll} 
        \hat{E}_k^{t}+\gamma_{1},   & if~\hat{E}_k^t~is~in~start~stage\\ 
        \hat{E}_k^{t}+\gamma_{2},   & if~\hat{E}_k^t~is~in~arise~stage\\
        \frac{1}{2}\hat{E}_k^{t},   & if~client~k~drops~out
    \end{array}
    \right.
    \end{split}
    \label{eqa:Fassa}
\end{equation}


\begin{algorithm}[t]
    \setcounter{algorithm}{2}
	\caption{FedSAE-Fassa: Federated Self-Adaptive Epoch with fastly starting and slowly arising (Proposed Framework)}
	\label{alg:fedsae-fassa}
	\begin{algorithmic}[1]
	\footnotesize
	    \Require the actually affordable local epoch $\tilde{E}_{k}^{t}$ of client $\bm{k}$ at round t, the server-predict workloads $\hat{E}_{k}^{t}$, the easy workloads $\hat{L}_{k}^{t}$ , the difficult workloads $\hat{H}_{k}^{t}$.
	    \Ensure the predicted easy workloads $\hat{L}_{k}^{t+1}$,  the predicted difficult workloads $\hat{H}_{k}^{t+1}$.
		\Procedure {Server}{}
		\State Initialize $\bm{w}^0$, $\hat{L}_{k}^{0}$, $\hat{H}_{k}^{0}$, $\alpha$
		\For{each communication round $t=1,2,...,T$}
			\State $S_t \leftarrow$ Select $K$ clients according to selected probabilities.
			\State Server broadcasts $\bm{w}_t$ to all selected clients.
			\State $( \hat{L}_{k}^{t}, \hat{H}_{k}^{t}, \tilde{E}_{k}^{t} ) \leftarrow$ Read history data from $S_t$.
			\For{each client $k \in S_t$ in parallel do}
			    \State ${\theta}_{k}^t \leftarrow \alpha {\theta}_{k}^{t-1} + (1-\alpha)\tilde{E}_{k}^{t-1}$
			    \State $( \hat{L}_{k}^{t+1}, \hat{H}_{k}^{t+1}, \hat{E}_k^{t} ) \leftarrow$ \textbf{EpochPredict ($\hat{L}_{k}^{t}$, $\hat{H}_{k}^{t}$, $\tilde{E}_{k}^{t}$, ${\theta}_{k}^t$)}
			    
			//the function Client() is defined in Algorithm 2
				\State ${w}_k^{t+1}\leftarrow$~\textbf{Client($k$,~$\bm{w}^t$, $\hat{E}_k^{t}$)}
				
			\EndFor
			\State $\bm{w}^{t+1}\leftarrow \sum_{k=1}^{K} \frac{n_k}{n}\bm{w}_k^{t+1}$
		\EndFor
		\EndProcedure
		
		\Statex
		// run on server
		\Function{EpochPredict}{$\hat{L}_{k}^{t}$, $\hat{H}_{k}^{t}$, $\tilde{E}_{k}^{t}$, ${\theta}_{k}^t$}  
		
	// client $\bm{k}$ will not drop out.
		\If{ 
			the actual affordable workloads greater than $\hat{H}_{k}^{t}$}
		    \Statex~~~~~~~~//client is in the arise stage.
			\If{ ${\theta}_{k}^t$ equal or less than $\hat{L}_{k}^t$ }
			    \State $\hat{L}_{k}^{t+1}, \hat{H}_{k}^{t+1} = \hat{L}_{k}^{t} + {r}_{2}, \hat{H}_{k}^{t} + {r}_{2}$
			    
			\ElsIf{ ${\theta}_{k}^t$ in ($\hat{L}_{k}^t$,$\hat{H}_{k}^t$]}
			     \State $\hat{L}_{k}^{t+1}, \hat{H}_{k}^{t+1} = \hat{L}_{k}^{t} + {r}_{1}, \hat{H}_{k}^{t} + {r}_{2}$
			 \Statex~~~~~~~~//client is in the start stage.
			 \Else
			      \State $\hat{L}_{k}^{t+1}, \hat{H}_{k}^{t+1} = \hat{L}_{k}^{t} + {r}_{1}, \hat{H}_{k}^{t} + {r}_{1}$
	        \EndIf
			\State $\hat{E}_{k}^{t}  = \hat{H}_{k}^{t}$
			
			// client $\bm{k}$ will drop out but weight at epoch $\hat{L}_{k}^{t}$ will be uploaded.
		\ElsIf{$E_{k}^{t}$ in ($\hat{L}_{k}^{t}$,$\hat{H}_{k}^{t}$] }
			\If{ ${\theta}_{k}^t$ equal or greater than $\hat{L}_{k}^t$ }
			      \State $\hat{L}_{k}^{t+1} = min(\hat{L}_{k}^{t} + {r}_{2},  \frac{1}{2}\hat{L}_{k}^{t})$
			      \State $\hat{H}_{k}^{t+1} = max(\hat{L}_{k}^{t} + {r}_{2},  \frac{1}{2}\hat{H}_{k}^{t})$
			\ElsIf{ ${\theta}_{k}^t$ in ($\hat{L}_{k}^t$, $\hat{H}_{k}^t$] }
			      \State $\hat{L}_{k}^{t+1} = min(\hat{L}_{k}^{t} + {r}_{1},  \frac{1}{2}\hat{H}_{k}^{t})$
			      \State $\hat{H}_{k}^{t+1} = max(\hat{L}_{k}^{t} + {r}_{1},  \frac{1}{2}\hat{H}_{k}^{t})$
			 \Else
			      \State $\hat{L}_{k}^{t+1} = min(\hat{L}_{k}^{t} + {r}_{1},  \frac{1}{2}\hat{H}_{k}^{t})$
			      \State $\hat{H}_{k}^{t+1} = max(\hat{L}_{k}^{t} + {r}_{1},  \frac{1}{2}\hat{H}_{k}^{t})$
			 \EndIf
			      \State $\hat{E}_{k}^{t} = \hat{L}_{k}^{t}$
			    
			// client k will drop out and no weight will be uploaded.
	    \Else
			 \State $\hat{L}_{k}^{t+1}, \hat{H}_{k}^{t+1} = \frac{1}{2}\hat{L}_{k}^{t}, \frac{1}{2}\hat{H}_{k}^{t}$
			 \State $\hat{E}_{k}^{t} = 0 $ 
		\EndIf
		
		\State \Return $\hat{L}_{k}^{t+1}$, $\hat{H}_{k}^{t+1}$, $\hat{E}_k^{t}$ 
	    \EndFunction

		
				
				
	\end{algorithmic}
\end{algorithm}

\begin{figure}[t]
	\centering
	\includegraphics[width=8.8cm]{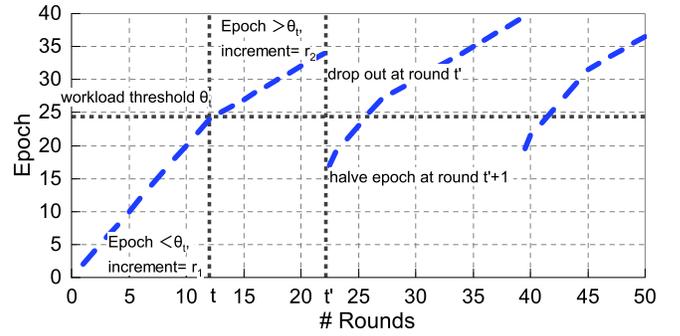}
	\vspace{-8mm}
	\caption{The simple process diagram of \textit{FedSAE-Fassa}. Each round clients calculate their workload threshold $\theta$ according to the historical information. In the start stage where the clients' local epoch less than $\theta$ and clients' local epoch increases by $\gamma_{1}$ each round. In the arise stage where the number of client’s local epoch is greater than $\theta$ and clients' local epoch increases by $\gamma_{2}$, where $\gamma_{1}$ $\textgreater$ $\gamma_{2}$. Also, if clients drop out then the local epoch of the next round will be halved.}
	\label{fig:Fassa}
\end{figure}

Similarly, $\hat{E}_k^{t}$ is the predicted workload of $k$-th client at round $t$, $\gamma_{1}$ and $\gamma_{2}$ ($\gamma_{1}$ $\textgreater$ $\gamma_{2}$) are the hyperparameters related to the increment. In \textit{FedSAE-Fassa}, $L_k^t$ and $H_k^t$ are predicted according to~\eqref{eqa:Fassa}.
The workload threshold $\theta_{k}^t$ reflects the weighted average of the affordable workload till $t$-th round. 
To accurately predict the affordable workload of clients, \textit{FedSAE-Fassa} divides the workload growth process into two stages based on the workload threshold----start stage and arise stage.
Among them, the workload in the start stage is less than $\theta_{k}^t$ and the probability of drop out is low so the workload grows fastly with an increment $\gamma_{1}$ while the workload in the arise stage is bigger than $\theta_{k}^t$ and the client is easy to drop out so the workload grows slowly with an increment $\gamma_{2}$. Obviously, $\gamma_{1} \textgreater \gamma_{2}$. The algorithm \textit{FedSAE-Fassa} using segmentation strategy is advantaged in the following points: 
(1) It takes full advantage of historical performance data of clients and dynamically adjusts the weights of old affordable workload in threshold computation to avoid abusing the out-of-date information.
(2) It increases the local epoch with two different growth rates, which predicts the local epoch of clients more cautiously.
We show the process of \textit{FedSAE-Fassa} in Fig. \ref{fig:Fassa}.

\textbf{Client selection with Active Learning.} 
Although the aforementioned workload prediction algorithms can reduce stragglers, the quality of the global model should be considered further as not all clients are active in the training process. Therefore, we combine active learning (AL)\cite{settles2009active} with \textit{FedSAE} to select clients with high training value for the global model to participate in training in order to obtain a high-quality global model. 
In general, some clients' local models own large loss, and focusing on optimizing these models will greatly improve the quality of the global model\cite{goetz2019active}. Hence, AL aims to select clients with larger training loss to participate in training. In our framework, the server firstly converts the cross-entropy loss of each client into the corresponding training value to measure the importance of the client. And then, the server converts the training value into selected probability and selects participants for each round according to the probabilities.
The training value of client $k$ at round $t$ is calculated as the following equation:
\begin{equation}
    \begin{split}
        v_k^{t+1}= \left \{ 
        \begin{array}{ll} 
            \sqrt{n_k}\bar{l}_k^{t},   & if~client~k~is~selected\\ 
            v_k^{t},   & otherwise
        \end{array}
        \right.
     \end{split}
     \label{eqa:vk}
\end{equation}
where $n_k$ is the number of samples on client $k$, $\bar{l}_k^t$ is the average loss of client $k$ at round $t$. 
At each round, only the loss of participants is refreshed. 
Each participant computes the evaluation of value $v_k^t$ and returns it to the server. 
Then the server turns $v_k^t$ into a selected probability $p_k^t$ by the equation as follows:
\begin{equation}    
    p_k^t=\frac{e^{\beta*v_k^t}}{\sum_{k}^{K}e^{\beta*v_k^t}}
    \label{eqa:pk}
\end{equation}
where $\beta$ is the hyper parameter to scale $v_k$ to prevent $e^{\beta*v_k}$ from being too large of too small to reflect the numerical relationship between the raw data $v_k$. 
Then the server picks participants out according to the selected probabilities $\{p_1, p_2,...,p_n\}$.

In terms of the cost of \textit{FedSAE}, compared to \textit{FedAvg}, the server requires $\mathcal{O}(n)$ extra time to calculate the affordable workload and selected probability. Besides, each client requires $\mathcal{O}(1)$ extra space to store the local loss and $\mathcal{O}(1)$ extra space to store the workload information of its latest model. However, the above-mentioned additional calculation overhead and storage overhead are small enough and can be ignored, which means our algorithm is practical.
To verify the rationality of our framework, we will introduce multiple experiments we conducted in the next section.

\section{evaluation}\label{sec:evaluation}
In this section, we present the experimental details and results for our framework \textit{FedSAE}.
In subsection \ref{subsec:affect-prediction}, we show the performance improvement of \textit{FedSAE} in heterogeneous systems. 
In subsection \ref{subsec:affect-AL}, we sufficiently exhibit the effects of \textit{FedSAE} combined with AL on convergence. 

\subsection{Experimental Details}\label{subsec:experimental details}
\begin{table}[b]
   \centering
   \caption{Statistics of FEMNIST, MNIST, Synthetic(1,1) and Sent140.}
\renewcommand\arraystretch{1.2}
\setlength{\tabcolsep}{5mm}{
\begin{tabular}{lccc}
\hline
Dataset                             & Model                                      & Devices & Samples \\ \hline
\multicolumn{1}{l|}{MNIST}        & \multicolumn{1}{c|}{\multirow{3}{*}{MCLR}} & 1,000   & 69,035  \\
\multicolumn{1}{l|}{FEMNIST}          & \multicolumn{1}{c|}{}                      & 200     & 18,345  \\
\multicolumn{1}{l|}{Synthetic(1,1)} & \multicolumn{1}{c|}{}                      & 100      & 75,349   \\ \cline{2-2}
\multicolumn{1}{l|}{Sent140}        & \multicolumn{1}{c|}{LSTM}                  & 772     & 40,783  \\ \hline
\end{tabular}
}
\label{tab:datasets}
\end{table}

\textbf{Datasets $\&$ Models.} Our evaluation includes two model families on four datasets which are consisted of both image classification tasks and text sentiment analysis tasks. 
\textbf{Federated Extended MNIST (FEMNIST)}\cite{caldas2018leaf} is a 26-class image classification dataset composed of handwritten digits and characters which partition the 62-class data in EMNIST\cite{cohen2017emnist} according to writers. 
The data of FEMNIST is distributed into 200 devices where each device holds only 5 classes as heterogeneous settings. 
We use multinomial logistic regression (MCLR) with 7850 parameters to train a convex model on FEMNIST. 
\textbf{MNIST}\cite{lecun1998gradient} is a 10-class image classification dataset composed of handwriting digits 0-9. 
On MNIST, 18,345 samples are divided into 1,000 devices. 
To simulate obvious heterogeneity, each device holds only 2-class digits and the samples of devices follow a power law. 
We optimize the model on MNIST with the same training model MCLR as FEMNIST. 
\textbf{Sentiment140 (Sent140)}\cite{go2009twitter} is a text sentiment analysis dataset composed of many tweets where each tweet can be extracted as a positive or negative sentiment. 
The number of clients on Sent140 is 772. 
We use LSTM to train the model on Sent140. 
\textbf{Synthetic}\cite{shamir2014communication} is a synthetic federated dataset that consists of 100 devices. 
We generated Synthetic dataset with the hyperparameter $\alpha=1$, $\beta=1$ (e.g. Synthetic(1,1)) and the data size on clients follows a power law. 
We use the MCLR classifier to train the model on Synthetic(1,1). 
The statistics of the above four datasets are shown in TABLE \ref{tab:datasets}.

\textbf{Federated Parameters.} 
All of the above models support training with at least one iteration. We split an epoch into multiple iterations. For example, if a client is required to train 3.5 epochs, then this client will train 3 epochs and  $0.5\tau$ iterations where $\tau=\frac{S_{k}}{B}$ varies with the practical iterations of an epoch, $S_{k}$ is the number of samples for client $k$, $B$ is the local mini-batch. 
In our experiments, we simulate systems heterogeneity by generating the affordable local workloads $E$ of clients at each round with Gaussian Distributions $E \sim \mathcal{N}(\mu,\sigma^{2})$ where $\mu$ $\in$ $[5,10)$, $\sigma \in [\frac{\mu}{4},\frac{\mu}{2})$, $\mu$, $\sigma$ are valued from their regions uniformly. 
We use the same federated learning settings as introduced in \ref{subsec:motivation}: the number of selected clients per round $K=30$ for MNIST, $K=10$ for FEMNIST, Sent140 and Synthetic(1,1), the mini-batch for local SGD is $B=10$, the learning rate $\eta$ for FEMNIST, MNIST, Sent140, Synthetic(1,1) respectively are 0.03, 0.03, 0.3, 0.01.  
Specific to \textit{FedSAE}, we initialize $(L_k^{0},H_k^{0})$ to (1,2) and we set the inverse ratio parameter $\mathcal{U}=10$, smooth index $\alpha=0.95$, the increment $\gamma_{1}=3$, $\gamma_{2}=1$, scale parameter of AL $\beta=0.01$. For the convenience of understanding, we will reiterate the meaning and value of above \textit{FedSAE} parameters again when we use it.

\begin{figure}[b]
	\centering
	\includegraphics[]{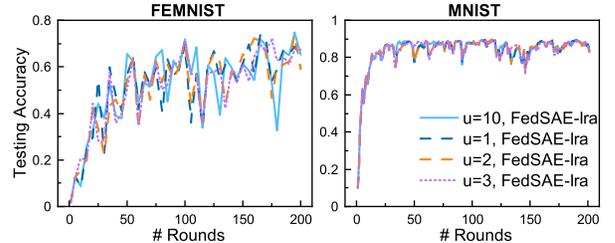}
	\vspace{-3mm}
	\caption{Effects of choosing different inverse ratio parameter $\mathcal{U}$. We show the experimental testing accuracy of global model on FEMNIST and MNIST datasets when $\mathcal{U} = 1, 2, 3, 10$. Empirically, we find that when $\mathcal{U} = 10$, \textit{FedSAE-Ira} works well.}
	\label{fig:Ira-choose-τ}
\end{figure}

\textbf{Baseline $\&$ Metrics.} The baseline of this paper is the vanilla FL framework  \textit{FedAvg}\cite{mcmahan2017communication}.
In our experiments, the affordable workload of each client changes over time. We fix the random seed to ensure that the same client has the same affordable workload set on different datasets. If the assignment of clients is less than its affordable workload, then the client can complete the training task, otherwise, it will drop out. For \textit{FedAvg}, the server fixes the assigned workload $E$ of clients to 15. For \textit{FedSAE}, the workload of clients is predicted by the server. In order to eliminate the influence of client selection, we fixed the random seed of each round to ensure that when the training framework is different, the server will select the same clients in the same round of the same dataset.
Each round we evaluate the global model. To compare each framework intuitively, we draw the top-1 testing accuracy, training loss of global model and drop out rate of clients in the following sections.

\begin{figure*}[t]
	\centering
	\includegraphics[scale=0.9]{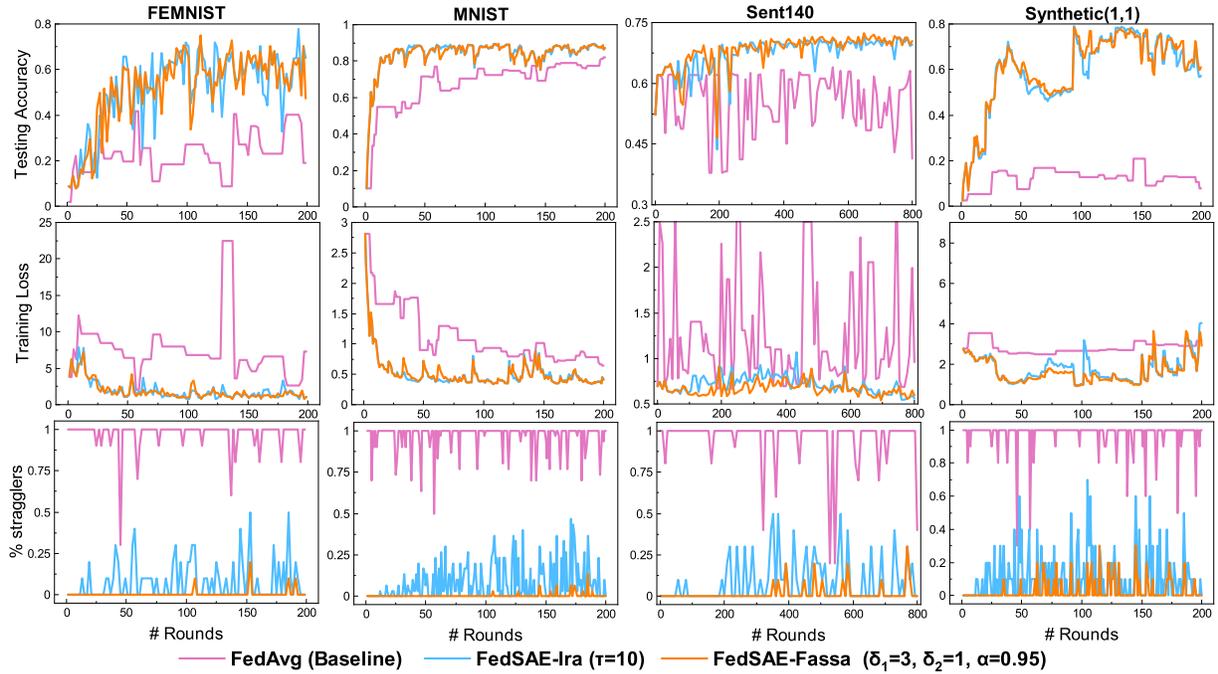}
	\vspace{-3mm}
	\caption{Testing accuracy, training loss, drop out rate of \textit{FedSAE-Ira}, \textit{FedSAE-Fassa} relative to \textit{FedAvg} on four datasets. We simulate a heterogeneous system by giving devices different parameters $\mu$, $\sigma$ of affordable workloads distribution function $\mathcal{N}(\mu, \sigma^2)$. \textit{FedSAE-Ira} and \textit{FedSAE-Fassa} can predict the affordable workloads of devices while \textit{FedAvg} can not. (1) \textbf{\textit{FedAvg vs. FedSAE-Ira.}} \textit{FedSAE-Ira} can increase testing accuracy up to 58\% and mitigate straggling up to 88.3\%. (2) \textbf{\textit{FedAvg vs. FedSAE-Fassa.}} Similar to \textit{FedSAE-Ira}, \textit{FedSAE-Fassa} brings testing accuracy improvement up to 57.5\% and mitigate straggling up to 96.3\%. (3) \textbf{\textit{FedSAE-Ira vs. FedSAE-Fassa.}} \textit{FedSAE-Ira} improves accuracy +0.4\% higher than \textit{FedSAE-Fassa} while \textit{FedSAE-Fassa} mitigates stragglers +6.875\% than \textit{FedSAE-Ira}. \textit{FedSAE} improves absolute testing accuracy by 26.7\%. We also report accuracy improvement in TABLE \ref{tab:fedavg-ira-fassa}.}
	\label{fig:fedavg-ira-fassa}
\end{figure*}
\begin{table*}[t]
	\centering
	\caption{Results of \textit{FedSAE-Ira, FedSAE-Fassa} on FEMNIST, MNIST, Sent140 and Synthetic(1,1). We show the top-1 testing accuracy and the average drop out rate( see the row of \% stragglers in the TABLE) as below. Besides, the percentages of accuracy improvement and stragglers mitigation are calculated.}
	\renewcommand\arraystretch{1.2}
	\setlength{\tabcolsep}{5mm}{
		\begin{tabular}{lcccccc}
			\hline
			\multicolumn{1}{c}{\multirow{2}{*}{Dasaset}} & \multicolumn{2}{c}{FedAvg (Baseline)}                                       & \multicolumn{2}{c}{FedSAE-Ira (Ours)}                             & \multicolumn{2}{c}{FedSAE-Fassa (Ours)}                           \\ \cline{2-7} 
			\multicolumn{1}{c}{}                         & \multicolumn{1}{l}{accuracy} & \multicolumn{1}{l}{\% stragglers} & \multicolumn{1}{l}{accuracy} & \multicolumn{1}{l}{\% stragglers} & \multicolumn{1}{l}{accuracy} & \multicolumn{1}{l}{\% stragglers} \\ \hline
			FEMNIST                                      & 41.7                         & 97.5                              & \textbf{77.8 ($\uparrow$ 36.1)}                         & 10.2 ($\downarrow$ 87.3)                             & 75.1 ($\uparrow$ 33.4)                         & \textbf{8.0 ($\downarrow$ 89.5)}                              \\
			MNIST                                        & 81.9                         & 96.6                              & \textbf{89.4 ($\uparrow$ 7.5)}                        & 8.3 ($\downarrow$ 88.3)                              & \textbf{89.4 ($\uparrow$ 7.5)}                        & \textbf{0.3 ($\downarrow$ 96.3)}                              \\
			Sent140                                      & 63.9                             & 96.5                                  & 72.1 ($\uparrow$ 8.2)                             & 10.3 ($\downarrow$ 86.2)                                  & \textbf{72.4 ($\uparrow$ 8.5)}                             & \textbf{1.4 ($\downarrow$ 95.1)}                                  \\
			Synthetic(1,1)                               & 20.9                         & 97.1                              & \textbf{78.9 ($\uparrow$ 58)}                        & 11.2 ($\downarrow$ 85.9)                             & 78.4 ($\uparrow$ 57.5)                        & \textbf{2.6 ($\downarrow$ 94.5)}                              \\ \hline
		\end{tabular}
	}
	\label{tab:fedavg-ira-fassa}
\end{table*}
\subsection{Effects of Affordable Workload Prediction}\label{subsec:affect-prediction}
We divide the experiment of \textit{FedSAE} into three groups: \textit{FedAvg vs. FedSAE-Ira}, \textit{FedAvg vs. FedSAE-Fassa}, \textit{FedSAE-Ira vs. FedSAE-Fassa}. The former two groups are to illustrate the effects of \textit{FedSAE}, and the latter group is to illustrate the performance difference between the two prediction algorithms. The details are following:

\begin{figure}[t]
	\centering
	\includegraphics[]{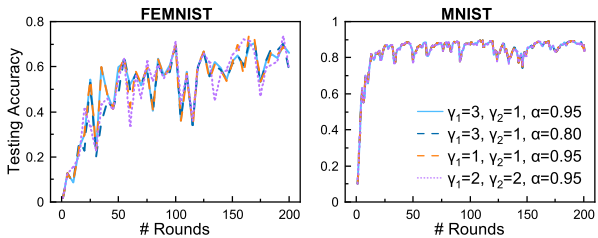}
	\vspace{-3mm}
	\caption{Effects of assigning different values to $\gamma_1$, $\gamma_2$ and $\alpha$ in \textit{FedSAE-Fassa}. We see that when $\gamma_1=3$, $\gamma_2=1$, $\alpha=0.95$, \textit{FedSAE-Fassa} works well on both FEMNIST and MNIST.}
	\label{fig:Fassa-choose-t}
\end{figure}
\textbf{\textit{FedAvg vs. FedSAE-Ira.}} \textit{FedSAE-Ira} predicts clients' local epoch according to the history of completing tasks of the latest round.
In \textit{FedSAE-Ira}, we predict the affordable workload with the increment which is inversely proportional to the amount of the client's current work. 
We conduct several experiments on FEMNIST and MNIST to get a suitable value of the inverse ratio parameter $\mathcal{U}$, and some details are shown in Fig. \ref{fig:Ira-choose-τ}. 
Empirically, we set $\mathcal{U}$ to 10 so the increment is $\frac{10}{E_k^t}$ if the client performs its assigned workloads successfully. 
To eliminate the influence of client selection, the server selects clients randomly on both \textit{FedSAE-Ira} and \textit{FedAvg}.
We implement \textit{FedSAE-Ira} (Algorithm \ref{alg:fedsae-ira}) in Tensorflow\cite{abadi2016tensorflow} and report the performance of \textit{FedSAE-Ira} and \textit{FedAvg} in Fig. \ref{fig:fedavg-ira-fassa} and TABLE \ref{tab:fedavg-ira-fassa}. 
We can see that on the four datasets, systems heterogeneity leads to more than $90\%$ of stragglers (see the pink solid lint in the bottom row of Fig. \ref{fig:fedavg-ira-fassa}). Moreover, from the training loss in the second row of Fig. \ref{fig:fedavg-ira-fassa}, we can see that \textit{FedAvg} has a divergent trend on FEMNIST, Sent140 and Synthetic(1,1). The evidence is that the training loss does not decrease but increases. In such a situation, \textit{FedSAE-Ira} effectively improves the testing accuracy by up to $58\%$ (see the line of Synthetic(1,1) in TABLE \ref{tab:fedavg-ira-fassa}), and reduces the stragglers by up to $88.3\%$. Besides, the convergence rate of the global model is restored. Because \textit{FedSAE-Ira} predicts an appropriate workload for each client, thus a large number of clients avoid dropping out.

\textbf{\textit{FedAvg vs. FedSAE-Fassa.}} \textit{FedSAE-Fassa} predicts clients' local epoch according to the training history of all of the past communication rounds. 
We show the experimental results of choosing the suitable parameters (i.e. $\alpha$, $\gamma_{1}$, $\gamma_{2}$) for \textit{FedSAE-Fassa} in Fig. \ref{fig:Fassa-choose-t}. 
Empirically, we set smooth index $\alpha$ to 0.95.  
And we set the increment parameters $\gamma_{1}$, $\gamma_{2}$ to $3$, $1$. 
The participants of each round also are selected randomly. 
The experimental results are shown in Fig. \ref{fig:fedavg-ira-fassa}. 
Similar to \textit{FedSAE-Ira}, \textit{FedSAE-Fassa} brings the accuracy improvement and mitigates straggling.
Specifically, \textit{FedSAE-Fassa} improves the testing accuracy by up to $57.5\%$, which is similar to \textit{FedSAE-Ira}. Besides, \textit{FedSAE-Fassa} also reduces stragglers up to $96.3\%$, which is $8\%$ higher than \textit{FedSAE-Ira}. The reason may be that \textit{FedSAE-Fassa} makes full use of clients' past training history of completing tasks, which can more accurately predict the changing characteristics of the affordable workload.

\textbf{\textit{FedSAE-Ira vs. FedSAE-Fassa.}} In Fig. \ref{fig:fedavg-ira-fassa},  \textit{FedSAE-Ira} and \textit{FedSAE-Fassa} are efficient in accuracy improving and straggler decreasing. 
To be exact, \textit{FedSAE-Ira} brings accuracy improvement +0.4\% higher than \textit{FedSAE-Fassa} on average. While \textit{FedSAE-Fassa} mitigates straggling +6.875\% more significantly than \textit{FedSAE-Ira} on average. From the data in TABLE \ref{tab:fedavg-ira-fassa}, we see that our framework \textit{FedSAE} improves absolute testing accuracy by 26.7\% and reduces the straggle rate by 90.3\% on average.

Although accuracy is improved, the convergence rate is still changeless. 
So we import Active Learning to accelerate the model convergence rate as described in the following section.
\begin{figure}[t]
	\centering
	\includegraphics[scale=1.0]{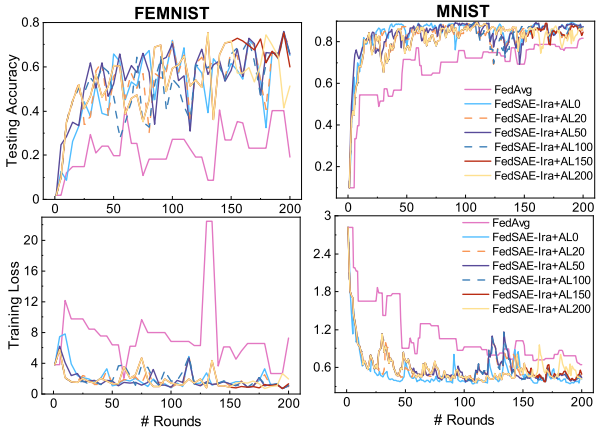}
	\vspace{-3mm}
	\caption{Testing accuracy and training loss on \textit{FedSAE-Ira} with the different training rounds using AL. \textit{FedSAE-Ira}+AL200 also named full \textit{FedSAE-Ira}+AL owns the fastest convergence rate.}
	\label{fig:Ira-AFL0-200}
\end{figure}
\begin{table}[t]
	\centering
	\caption{The number of training rounds for FEMNIST to achieve the goal testing accuracy 60\% and MNIST to achieve the goal testing accuracy 84\%.}
	\renewcommand\arraystretch{1.2}
	\setlength{\tabcolsep}{1.9mm}{
		\begin{tabular}{lccccccl}
			\hline
			\multirow{2}{*}{Dataset} & \multicolumn{6}{l}{\# Rounds of using AL in FedSAE-Ira} & \multirow{2}{*}{\begin{tabular}[c]{@{}l@{}}FedAvg,\\ E=15\end{tabular}} \\ \cline{2-7}
			& 0       & 20     & 50     & 100     & 150     & 200     &                                                                         \\ \hline
			FEMNIST                  & 48      & 30     & 31     & 37      & 32      & 32      & \multicolumn{1}{c}{-}                                                   \\
			MNIST                    & 25      & 19     & 21     & 19      & 19      & 19      & \multicolumn{1}{c}{-}                                                   \\ \hline
		\end{tabular}
	}
	\label{tab:AFL-convergence}
\end{table}

\subsection{Effects of AL}\label{subsec:affect-AL}  
To improve the quality of the global model further, we utilize Active Learning (AL)\cite{settles2009active} to dynamically select participants with high training value for the global model. 
Empirically, we set the scale parameter $\beta$ to $0.01$ which is the same as \cite{goetz2019active}. 
As shown in Fig. \ref{fig:Ira-AFL0-200}, we conduct multiple experiments with the number of rounds to perform AL as a variable, where the label \textit{FedSAE-Ira}+ALn represents the client selection strategy is used in the first n training rounds. Particularly, \textit{FedSAE-Ira}+AL0 represents the pure \textit{FedSAE-Ira}. Similarly, \textit{FedSAE-Ira}+AL200 represents full \textit{FedSAE-Ira}+AL. The experimental results are shown in Fig. \ref{fig:Ira-AFL0-200} which demonstrates that AL improves the convergence speed of the global model. In order to quantify the impact of AL on convergence, we show the number of training rounds for FEMNIST to achieve the goal accuracy 60\% and for MNIST to achieve the goal accuracy 84\% in TABLE \ref{tab:AFL-convergence}. 
From TABLE \ref{tab:AFL-convergence}, we can see that in the early of the training, compared to pure \textit{FedSAE-Ira} and \textit{FedAvg}, \textit{FedSAE-Ira+AL} can increase the convergence speed of the model by at least 23.5 \% on average. However, AL improves the convergence speed while slightly reducing the testing accuracy of the global model i.e. accuracy of FEMNIST and MNIST decreases by up to 2.3\%, 0.7\% separately. To achieve a compromise between convergence rate and global model accuracy, we recommend using the client selection strategy for the first quarter of the training round.

\section{conclusion}\label{sec:conclusion} 
In this paper, we propose \textit{FedSAE}, aiming to solve the straggle problems and performance reduction caused by systems heterogeneity. 
\textit{FedSAE} allows clients to self-adaptively perform their affordable workloads according to the training history of clients which relies on the two self-adaptive algorithms \textit{FedSAE-Ira} and \textit{FedSAE-Fassa}. 
We explain the design details of these two algorithms and parameters choosing to get efficient performance. 
Also, we show a series of experimental results on federated datasets to demonstrate that the significant accuracy improvement and stragglers decreasing of our framework in heterogeneous federated learning systems.
Our framework, \textit{FedSAE}, generally improves 26.7\% absolute testing accuracy and reduces 90.3\% drop out rate on average. Besides, \textit{FedSAE} can speed up convergence rate of global model.
Our future work includes providing convergence analysis for our framework and providing more reliable accuracy improvements for imbalanced datasets when we accelerate model convergence.

\bibliographystyle{IEEEtran}

\bibliographystyle{IEEEtran}
\balance
\bibliography{IEEEabrv,references}



\end{document}